# Engineering an Intelligent Essay Scoring and Feedback System: An Experience Report

Akriti Chadda, Kelly Song, Raman Chandrasekar, Ian Gorton
Khoury College Of Computer Sciences
Northeastern University
Seattle, USA
{chadda.ak, song.ke, r.chandrasekar, i.gorton}@northeastern.edu

*Abstract*— Artificial Intelligence (AI) / Machine Learning (ML)-based systems are widely sought-after commercial solutions that can automate and augment core business services. Intelligent systems can improve the quality of services offered and support scalability through automation. In this paper we describe our experience in engineering an exploratory system for assessing the quality of essays supplied by customers of a specialized recruitment support service. The problem domain is challenging because the open-ended customer-supplied source text has considerable scope for ambiguity and error, making models for analysis hard to build. There is also a need to incorporate specialized business domain knowledge into the intelligent processing systems. To address these challenges, we experimented with and exploited a number of cloud-based machine learning models and composed them into an application-specific processing pipeline. This design allows for modification of the underlying algorithms as more data and improved techniques become available. We describe our design, and the main challenges we faced, namely keeping a check on the quality control of the models, testing the software and deploying the computationally expensive ML models on the cloud.

## I. Introduction

Recent advancements in AI and ML, the availability of open-source tools and libraries, and the decreasing costs of hosting applications on the cloud have created opportunities for small and big businesses alike [1]. However, the engineering challenges surrounding AI/ML based applications prove to be significant [11], especially when a transition from prototype to production occurs [12]. This paper describes our work in engineering a prototype intelligent business system with the aim to make evolution and transition to production as frictionless as possible. We describe our experience in translating complex business requirements from our client into an initial version of a cloud-based ML application. Adding automated, intelligent processing is fundamental to the future scalability and hence success of our client.

Some of the biggest challenges faced by developers for small companies embarking on ML projects are following an experimental process, the difficulty of identifying customer business metrics and designing the database architecture [2]. To this end, we loosely followed the development process described by Hill et al [13]. This involves iterative activities centered on problem definition and refinement, data collection, algorithm and feature selection, and model evaluation.

In the following sections, we describe the business requirements and their realization into a cloud-based software framework. We describe the solution architecture adopted, and the challenges faced in deployment – both technical and managerial.

We also recognized the looming complexities of managing source and derived data, metadata and other common artefacts [14]. To this end, we decoupled the design and evolution of the ML and database subsystems through a well-defined interface that supports interactions between the two. The detailed database design is however beyond the scope of this paper.

## II. Business Problem Requirements

Our client is a small business with a rapidly growing customer base. The business allows customers to send in recruitment-related essays which are then graded by the employees on a number of dimensions based on a predetermined rubric. The scores 0 (unsatisfactory), 1 (improvements needed) or 2 (satisfactory) are given for each of the dimensions and then a weighted average of these scores is provided. This score is taken as a proxy for the quality and completeness of the knowledge and sentiments expressed in the essay.

In addition, textual feedback is given for each of the dimensions. This contains both constructive comments when more work is needed as well as appreciation for a point that is well written.

Our client came to us with a set of primary requirements, namely:

- the ability to streamline the collection of data from customers,
- automatic scoring of essays based on a set of rubrics,
- automatic feedback generation for different sections of the essays,
- generation of a final report for the customer,
- managing the data and process in a persistent store, and
- a cloud-based deployment

In essence, the requirements revolve around replacing the expert knowledge of the essay scorers with an intelligent software system [15][16]. From a machine learning perspective, this is a challenging scientific problem, requiring significant domain knowledge to be embedded in machine learning models.

Hence, given the time-boxed nature of this initial project engagement, we agreed with the client to take an architectural approach to design and development. The resulting architectural framework will then serve as a foundation for experimentation with the models to improve quality of outputs over time.

To this end, the focus of our work was:

- creating an extensible, modifiable, scalable cloud-based framework for the application
- developing initial ML models for the various dimensions of essay scoring, deployed within the framework.

For model development, we were provided with a dataset consisting of 1000 essays from the business' existing customers. Each of these had been manually scored and hence individual scores per dimension, aggregated scores and textual feedback was available for training purposes. There were 13 dimensions for which scores had to be calculated. This meant 13 different ML models would be needed, each trained to provide insights and feedback on a single dimension.

From an ML perspective, 1000 documents is a small data set for training purposes [17, 18] This is especially true for problems involving language models, where every word could be a feature, leading to high-variance issues; more data helps in reducing overfitting. Hence, we did not at this stage of the development have access to enough training data to build highly effective models. It was therefore a requirement for our system to evolve downstream as larger volumes of training data became available. This means the models should be capable of either online training or provide a straightforward mechanism to update the models when subsequent offline training had been performed.

III. SOFTWARE ARCHITECTURE

The application processing pipeline follows three logical phases, which are illustrated in Figure 1.

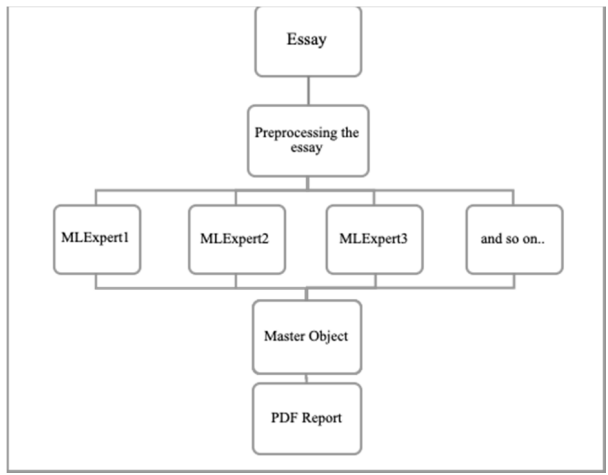

Fig. 1. Flow of control through the system, starting with the preprocessing of the input essay, followed by the data going to all the experts in parallel, and then a "MasterObject" for the experts to store the final scores and generate feedback.

**Preprocessing of essay text:** The first step is to normalize the client-provided textual inputs for the downstream ML algorithms. This preprocessing includes taking care of punctuations, performing named entity recognition for certain kinds of words, removal of unnecessary stop words and formatting of the input to lower case. The output of this preprocessing is transformed into a custom datatype and then passed onto the next phase of processing.

**Essay processing:** A copy of the normalized essay is passed to multiple decoupled *MLExpert,* which operate in parallel to process a particular essay dimension. This architecture exploits the fact that dimensions are orthogonal and require a specific ML model each to provide the necessary automation. It also supports modifiability in that models can be independently updated, and new models can be added if new dimensions are added to the business problem.

**Results aggregation:** Each model stores its outputs, namely a score and textual feedback, in a *MasterObject* abstraction. This object has elements for the outputs of each model. We built a simple user interface to display the suggested scores for each dimension as well as the final cumulative score. The user interface incorporates templates to enable the machine-generated comments to be edited by essay assessors. Finally, once all the scores and feedback are approved, the application generates a PDF file with the details about the customer, scores and feedback that can be sent as a report to the customer.

Algorithmically, we decided to address the ML problem as a classification problem. Open-ended free text used in the input essays is hard to analyze since there is considerable scope for ambiguity, and therefore error, in the text. We therefore tried a variety of models including deep learning methods and more classical machine learning methods.

Even though the amount of training data was small, for some dimensions, we first tried using two popular word embeddings, BERT [19] and ELMo [20, 10]. These were used with simple neural network classifiers to understand the difference in evaluation of the models. For other dimensions we tried the Word2Vec [21] representation as well as different variants of BERT. The idea of these methods is not to use individual words as such, but to take into account the context of the word, so as to get better representations of each word, and hence of segments of text. Such language models come pre-trained on large text corpora, meaning these learned models can directly be applied to a variety of text problems which require rich representations. However, this comes with a disadvantage: the language models tend to be huge and pose challenges in deployment, as we describe below.

In addition, we developed and evaluated classifiers for all dimensions based on standard ML methods including Multinomial Bayes, and Support Vector Machines (SVM). In these systems, we used a much simpler Bag of Words (BOW) representation. Thus these models did not need any large language models and were more resource efficient. We fine-tuned the models using a grid search over a number of hyper-parameters.

Each model was trained and saved offline, significantly reducing the run time of each model. The deployed models used text classification algorithms using statistical methods for calculating the final results, along with a Support Vector Machines based on the Bag Of Words representation.

We also built the system so the basic architecture could be executed in two modes. Local invocation from a command line supported testing and evaluation purposes. This mode supported the development team and was especially important for model refinement. The architecture could also be deployed on a cloud platform to facilitate operations and integrate with the data management system we built.

## IV. Cloud Deployment

The cloud infrastructure we created to run the application is hosted on AWS (Amazon Web Services). It was designed to accommodate the size and processing time of the ML/AI models used. The final cloud architecture (see Figure 2) is contained in a single Virtual Private Cloud (VPC) and consists of an Elastic Load Balancer (ELB), AWS Lambda, Elastic File System (EFS), and Elastic Compute Cloud (EC2) instances that execute models.

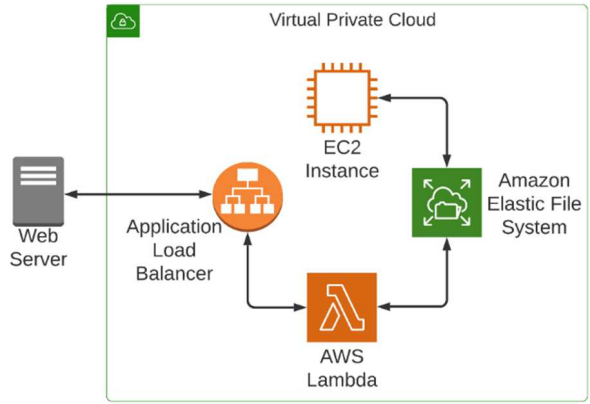

Fig. 2. Overview of Cloud infrastructure for the application

The system is invoked by sending the input essay as a web request, which triggers the application code housed in a AWS lambda function. The application dependencies, which are stored in the EC2 instance, are then accessed through the EFS access point, and the models, which reside on a collection of EC2 instances, are invoked.

## V. Challenges

Certain AI and cloud specific challenges arose during the design and construction of the application. These ultimately influenced the final architecture, as we discuss in the following.

We strived for strong modularity through the use of abstract classes and interfaces. The intent was that all of our *MLExperts* could be uniformly abstracted and implement the same interfaces. However, since the models we deployed were highly heterogeneous - built by different people on different operating systems with different dependencies - bringing them all together during deployment exposed us to the challenge of package versioning. This is an ongoing problem in the industry, especially because of the lack of complex logging mechanisms for AI systems. [3]. Due to the short project duration, tracking of our versions was done via informal methods, such as Slack and Github, with most of us keeping configurations locally until the model was trained and evaluated. As the project evolves, we need to address this issue with rigorous, repeatable approaches.

We also had issues with scalability. Developing large-sized models was relatively straightforward on our powerful development machines. We soon discovered deploying the same models on the cloud incurred high hosting costs to ensure adequate performance. Thus, to build an efficient cloud-based AI system, developers need to be aware of the difference between the development and deployment environments, and design the system accordingly to create an efficient and cost-effective ML/AI application. [4]

Testing of ML/AI based systems is challenging. Testing includes the verification of software to see if it behaves as expected. However, performing tests on ML systems is not straightforward because the procedures to perform the tests are different from the traditional techniques applied in non-ML systems. Traditional testing techniques aim to increase the coverage towards the exploration of diverse software states. [6] However, the understanding about errors found in ML models is currently limited, thus leading to arbitrary bug detection and repair approaches. [7] Owing to the amount of data and resources available in the industry, it is possible to employ combinatorial tests [8] and stress tests to measure the end-to-end performance of ML/AI web-based services. [9]. Exploiting these strategies is however not a straightforward exercise.

Additionally, interpretation of the model results is not always clear or straight forward. This is especially true when working with text data and using contextual word embeddings to vectorize the data. As mentioned earlier, we built and evaluated classifiers based on both rich representations such as ElMO and BERT, and using more standard ML methods on simple BOW representations.

For a variety of reasons, we did not choose these more complex models over the BOW models. Firstly, it was difficult to efficiently deploy these resource-hungry models without increasing the computing power and system cost significantly. Secondly, the difference in precision in the two models was insignificant, and hence insufficient to warrant these costs. Our test cases were admittedly limited. With textual classification problems, the input domain is essentially infinite and hence we need to design much deeper testing and evaluation strategies for our model downstream.

In addition, the long execution times of our models caused another architecture modification. We at first deployed an API Gateway, which acted as the client access point to our application business logic. However, the processing times of some of the ML models exceeded the 30 second timeout time of the API Gateway. We avoided this timeout by replacing the API Gateway with an Application Load Balancer, which will help with performance in the long term.

A final challenge was due to the cloud components we selected for implementing the architecture. Originally, we pursued a pure serverless compute infrastructure due to its cost efficiency and low ongoing maintenance requirements. To this end, we wanted to contain the application code in its entirety in a single AWS Lambda function. However, this proved to be impossible, as the size of the classifiers and long list of model dependencies vastly exceeded the maximum allowed deployment package size for a Lambda function.

Thus, we added an EFS instance as an application specific access point in tandem with an EC2 instance. The Lambda function still contained the ML pipeline code but was able to access the dependencies via the EFS access point to bypass the Lambda storage issue.

## VI. Conclusion

In our experience from this project, the transition from open-ended client requirements to an initial implementation of an intelligent system is not straightforward to achieve. In our case, the client wanted automation of the scoring of customer-submitted essays. To achieve this aim, we designed a modular software architecture that contains components for deploying and executing ML models. The individual ML model results are then combined together to get the final result. In this way, it is much easier to remove any unwanted models, change and evolve the algorithms of the existing models as well as add new criteria, without affecting how the application runs.

The development of this application is still in its infancy. The initial system we have built is being used as the basis for further developing the application capabilities and adding usability features. Considerable further model validation and enhancement is needed before all the dimensions of essay scoring can be adequately automated. However, we hope that a first operational version that can add busines value can be deployed later in 2021.

## VII. Acknowledgements

We acknowledge our deep gratitude to our client who brought us this interesting problem and participated actively in the development of this system. The client remains anonymous for commercial and confidentiality reasons. We are also extremely grateful to Chetna Khanna, Gabriel Rada, Matthew Gates-Dehn, Nicholas Nemetz, Philip Butler, Roopa Uma Shiv and Shucheng Chao from Khoury College of Computer Sciences for their input and work towards the application.